\documentclass[letterpaper, conference, 10pt,twocolumn ]{IEEEtran}
\pdfoutput=1
\let\chapter\section
\usepackage[ruled,vlined,linesnumbered]{algorithm2e}
\usepackage{graphicx}
\usepackage{amsfonts}
\usepackage{amsmath,amssymb,mathrsfs}
\usepackage{array}
\usepackage{flafter}
\usepackage{cite}
\usepackage[tight]{subfigure}
\usepackage{verbatim}
\usepackage{caption}
\usepackage{subfig}
\usepackage{color}
\usepackage[numbers]{natbib}
\usepackage{bm}
\usepackage{subfigure}
\usepackage{url}

\newtheorem{assumption} {Assumption}
\newtheorem{proposition} {Proposition}
\newtheorem{remark}{Remark}

\newtheorem{corollary}{Corollary}

\IEEEoverridecommandlockouts

\begin{document}
%
\title{Motion Planning in Non-Gaussian Belief Spaces (M3P): The Case of a Kidnapped Robot}

\author{\IEEEauthorblockN{Saurav Agarwal}
\and
\IEEEauthorblockN{Amirhossein Tamjidi}
\and
\IEEEauthorblockN{Suman Chakravorty}
\thanks{Saurav Agarwal (\texttt{sauravag@tamu.edu}), Amirhossein Tamjidi (\texttt{ahtamjidi@tamu.edu}) and Suman Chakravorty (\texttt{schakrav@tamu.edu}) are with the Department of Aerospace Engineering, Texas A\&M University, College Station, TX 77840.}}

\maketitle

\begin{abstract}
Planning under uncertainty is a key requirement for physical systems due to the noisy nature of actuators and sensors. Using a belief space approach, planning solutions tend to generate actions that result in information seeking behavior which reduce state uncertainty. While recent work has dealt with planning for Gaussian beliefs, for many cases, a multi-modal belief is a more accurate representation of the underlying belief. This is particularly true in environments with information symmetry that cause uncertain data associations which naturally lead to a multi-modal hypothesis on the state. Thus, a planner cannot simply base actions on the most-likely state. We propose an algorithm that uses a Receding Horizon Planning approach to plan actions that sequentially disambiguate the multi-modal belief to a uni-modal Gaussian and achieve tight localization on the true state, called a Multi-Modal Motion Planner (M3P). By combining a Gaussian sampling-based belief space planner with M3P, and introducing a switching behavior in the planner and belief representation, we present a holistic end-to-end solution for the belief space planning problem. Simulation results for a 2D ground robot navigation problem are presented that demonstrate our method's performance. 
\end{abstract}

\IEEEpeerreviewmaketitle

\section{Introduction}

Motion planning for robotics involves dealing with the uncertain nature of physical systems i.e. noisy actuators and sensors as well as changes in the environment in which the robot operates. The motion or actuator uncertainty makes it difficult to execute precise actions and sensing uncertainty makes it impossible to determine the exact state of the robot. Further, changes in the environment can reduce the effectiveness of plans computed offline. Thus, unless a plan can be updated on the fly to account for new constraints, the plan might fail. A significant amount of research has gone into developing probabilistic methods to achieve robust performance for practical systems. In the probabilistic approach, the aim is to develop methods that maximize the probability of achieving a desired state. State of the art methods rely on a probability distribution over the system's state (called the belief) and develop solutions in the belief space that enable us to drive the system belief from an initial to a desired belief. In the case of a mobile robot, we may wish to drive the system from a start location to a goal location, or in the case of a manipulator, manipulate an object from an initial to some final state. In general, planning for systems under uncertainty belongs to the class of Partially-Observable Markov Decision Process (POMDP) problems which are known to be computationally intractable. \\

Recent work, in particular, sampling based methods, have shown varying degrees of success in solving the POMDP problem. In \cite{Prentice09}, the authors construct a belief space variant of a Probabilistic RoadMap (PRM) \cite{Kavraki96} to generate plans that result in information gathering actions which minimize state uncertainty at the goal. The authors use linearized process and measurement models, and a Gaussian belief parametrization. In \cite{Bry11}, a graph is constructed in belief space and pruned successively, ultimately resulting in a tree in belief space which guarantees convergence to the optimal solution in the limit of infinite samples. These methods along with others \cite{chaudhari-ACC13}, \cite{Berg10}, \cite{kurniawati2012global} provide solutions that are dependent on the initial belief. This makes them computationally inefficient for real-time replanning. Feedback-based Information RoadMap (FIRM) \cite{Ali14-IJRR} also builds a belief space variant of a PRM, but this work differs from the previous ones in that it enables re-use of the offline computations to compute new policies online. FIRM introduces belief stabilizers at the graph nodes which act as funnels and lead to edge independence in the FIRM graph. We note that the different instantiations of this method also assume a Gaussian belief representation. This method is shown to work well in practice in \cite{Ali14-RolloutFIRM-ICRA} on a mobile robot operating indoors. \\



 \begin{figure}[h!]
  \centering
 {\includegraphics[width=1.8in]{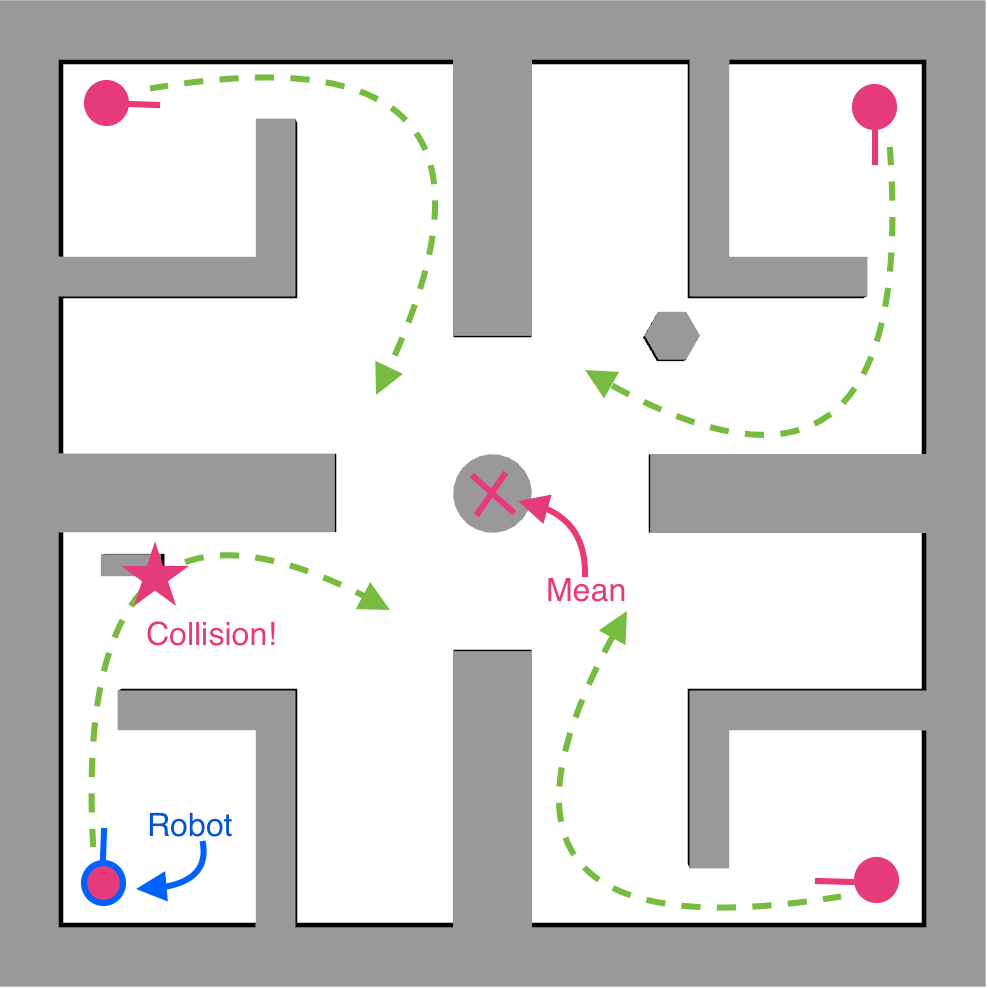}}
  \caption{A kidnapped robot scenario in a world with 4 identical rooms. The actual robot is depicted by the blue outlined disk, whereas the hypothesis are depicted by red disks. All hypothesis are equally likely. The dashed green arrows show a possible control action based on the hypothesis in the top-right room that can result in collision for the actual robot.}
  \label{fig:multi-modal-example}
\end{figure}

\subsection{Issues with Gaussian Belief Representation}


In \cite{Ali14-RolloutFIRM-ICRA} situations such as kidnapping (lost-robot), were dealt with by expanding the error covariance such that the Bayesian estimator (in this case a Kalman filter) is able to account for the unexpected data associations/observations or large innovations. This works well as long as there are no uncertain data associations. The robot was able to enter into an Information Gathering Mode and make small cautious movements to gain enough information until the error covariance converged to a small threshold. Certain situations can arise where data association between what is observed and the robot's a-priori map leads to a multi-modal hypothesis.
An example is the \textbf{data association problem} for a kidnapped robot operating in a symmetric environment. Imagine a mobile robot equipped with a laser scanner operating in a world where there are identical rooms as shown in Fig. \ref{fig:multi-modal-example}. To the laser, each and every room appears identical. Thus, if the robot is switched off and placed randomly in one of the rooms, on waking up it has no way of knowing exactly where it is. In a sense, based on the observations, the robot would think it could be in either of the four rooms at that instant since the laser readings would look alike. We can extend the previous statement to say that when sensory information leads to uncertain data associations, the robot may believe itself to be in one of multiple states at the same time. Such a situation implies that the pdf of the robot's belief cannot be represented by a uni-modal Gaussian. \\

In \cite{Platt11-isrr}, the authors investigate the grasping problem with a multi-modal hypothesis of the gripper's state. Their method picks the most-likely hypothesis and a fixed number of samples from the belief distribution, then using a Receding Horizon Control approach, belief space trajectories are found using direct transcription that maximize the observation gap between the most-likely hypothesis and the drawn samples. The authors use this approach to prove/disprove the most likely hypothesis.
In \cite{platt-correctness-icra12}, the correctness and complexity of the algorithm presented in \cite{Platt11-isrr} is analyzed and proven to eventually converge to the goal region in belief space with as low as two samples. In \cite{platt-wafr12-RHC}, the authors build upon \cite{Platt11-isrr} wherein they transpose the non-convex trajectory planning problem in belief space to a convex problem that minimizes the average log-weight of the belief-modes and a quadratic action cost. \\
Fundamentally, our work and the work in \cite{Platt11-isrr,platt-wafr12-RHC} aim to achieve the same goal of proving/disproving hypothesis of the robot's state. At the very basic level, these methods choose actions that lead to information gathering behavior such that different modes of the robot's belief are expected to observe different information, thus disambiguating the multi-modal hypothesis. 
In our method, we develop an information graph node based receding horizon planner that sequentially disambiguates the different hypothesis encoded in the multi-modal belief state. Further, the technique developed is guaranteed to drive the belief state into a unimodal belief state from where a Gaussian planner such as FIRM can take over to complete the robot's mission. We believe that our method is computationally more efficient than the optimization based technique in \cite{platt-wafr12-RHC, Platt11-isrr}. It is also able to deal with the kidnapped robot scenario which may not be possible to address using the trajectory optimization based technique in \cite{platt-wafr12-RHC, Platt11-isrr} due to the difficulty of generating an initial feasible plan for the widely separated modes in the presence of obstacles. We further believe that our planner is applicable to any situation wherein a multi-modal belief may arise in the robot's state due to uncertain data associations. We also show how the multi modal planner can be seamlessly combined with a unimodal belief space planner such as FIRM to facilitate belief space planning for robotic tasks where Gaussian/ unimodal belief representations may not be valid during the entirety of the task.

\textit{\textbf{Contributions}}: The key contributions of our work are as follows:

\begin{enumerate}
 \item We represent the belief with a Gaussian Mixture Model (GMM) rather than particles. 
 
 \item Instead of basing actions on the most-likely hypothesis, we create candidate actions based on each mode and evaluate the best one.
 
 \item We use a sampling based planner i.e. RRT* \cite{Karaman.Frazzoli:IJRR11} to plan candidate trajectories. (One can also simply use RRTs \cite{Lavalle01-RRT} but due to insignificant overhead in using RRT* over RRT we prefer RRT* as it gives us the benefit of optimality)
	
 \item We introduce a switching behavior in the belief representation during the online-phase from Gaussian to non-Gaussian, and back, as required. Our argument is that most of the times, the belief is well represented by a Gaussian, wherever this is not the case, we switch to a GMM and our algorithm creates plans that converge back to a uni-modal Gaussian.
  
 \item We present simulation results for a 2D navigation problem in which a robot is kidnapped.

 \end{enumerate}

\section{Problem statement}

Let $ x_{k} $, $ u_{k}$, and $ z_{k}$ represent the system state, control input, and observation at time step $ k $ respectively. Let $ \mathbb{X} $, $ \mathbb{U} $, and $ \mathbb{Z} $ denote the state, control, and observation spaces respectively. It should be noted that in our work, the state $x_k$ refers to the state of the mobile robot i.e. we do not model the environment and obstacles in it as part of the state. The sequence of observations and actions are represented as $ z_{i:j}=\{z_{i},z_{i+1},\cdots,z_{j}\} $ and $ u_{i:j}=\{u_{i},u_{i+1},\cdots,u_{j} \}$ respectively. The non-linear state evolution model $ f $ and measurement model $ h $ are denoted as  $x_{k+1}=f(x_{k},u_{k},w_{k})$ and $z_{k}=h(x_{k},v_{k})$, where $w_{k} \sim \mathcal{N}(0,Q_k)$ and $ v_{k} \sim \mathcal{N}(0,R_k)$ are zero-mean Gaussian process and measurement noise, respectively.\\

We assume that the robot is initially tasked to go from a start location to a goal location. For this purpose, we use a Gaussian Belief Space Planner such as FIRM. However, during the course of the task, say at time $k$, the robot is kidnapped, or gets lost, leading to uncertain data associations for its sensor readings. In such a case, 
the belief $b_k$ can be represented by a GMM at time $k$ as a weighted linear summation over Gaussian densities. Let $w_{i,k}$, $\mu_{i,k}$ and $\Sigma_{i,k}$ be the importance weight, mean vector and covariance matrix associated to the $i^{th}$ Gaussian $m_i$ respectively at time $k$, then

\begin{equation}\label{eq:gmm-linear-sum}
b_k = \sum_{i=1}^{N} w_{i,k} m_{i,k},  ~~ m_{i,k} \sim \mathcal{N}(\mu_{i,k}, \Sigma_{i,k}).
\end{equation}

\noindent\textit{ Our goal is to construct a belief space planner $\mu(b_k)$ such that under the belief space planner, given any initial multi-modal belief $b_0$, the belief state process evolves such that $b_T = m_T$, where $m_T = \mathcal{N}(\mu_T, \Sigma_T)$ for some finite time $T$.}\\

In other words, our goal is to construct a belief space planner such that it is guaranteed to drive the initial multi-modal belief into a unimodal belief in finite time.
The basic idea is that once such a unimodal belief is achieved, a Gaussian belief space planner such as FIRM can take over again and guide the robot towards its mission goal which was interrupted due to the kidnapping. \\

Note that we do not require optimality from our planner, only that it stabilize the belief state process to a unimodal belief in a guaranteed fashion. Further, albeit the scenario used to motivate the problem is the kidnapped robot situation, the method proposed is  general, and can be extended to any planning situation where a multi-modal belief arises in the robot state due to uncertain data associations.

\section{Methodology}

One cannot simply base actions on the mean, as in a multi-modal scenario, it would not make sense in the physical world. Again, taking the example shown in Fig. \ref{fig:multi-modal-example}, the mean of all the modes may lie within some obstacle which obviously is not a good choice to base actions on. Our algorithm creates candidate plans for each belief mode that guide it to a state called the target state such that the targets of all the modes have minimal information overlap. For each policy, we simulate the expected information gain which we define as the reduction in the discrete number of modes and choose the gain maximizing policy.


\subsection{Belief Propagation Using Gaussian Mixture Model}\label{sec:multi-hypothesis-representation}
We adopted the Gaussian-Mixture-Model (GMM) for belief representation because it provides a seamless transition from multiple to single hypothesis scenarios and vice versa during localization and planning. We found in our experiments that the particle filter alternative \cite{kwok2003adaptive} which is generally used for global localization in a kidnapped situation is slower in convergence and less robust compared to the approach presented here. Our method also incurs less computational cost compared to PF-Localization. 
Once the lost state is detected, we need to generate a rich set of samples from which we can converge to the most likely modes. We uniformly sample the configuration space and set these samples as the means $\mu_{i,k}$ of the modes of the Gaussian and assign equal covariance to each mode. Algorithm \ref{alg:sampling-gmm} shows how we generate the initial multi-modal belief. 
Note that $n_k$, the number of Gaussian-modes at time $k$, can vary depending on the observational sequences as new hypotheses are added and some die out. Further, we keep the weights normalized such that $\sum_{i=1}^{N} w_{i,k}=1$. As the robot moves and gets observations, the importance weights $w_{i,k}$'s are updated based on the measurement likelihood function as shown in Eq. \ref{eq:weight-update}.

\begin{equation}\label{eq:weight-update}
 w_{i,k+1} = \frac{w_{i,k} e^{-0.5D_{i,k}^{2}}}{\sum_{i=1}^N w_{i,k} e^{-0.5D_{i,k}^{2}}} 
\end{equation}

where $D_{i,k}$ is the Mahalanobis distance between the sensor observation and most-likely observation for mode $m_i$ such that

\begin{equation}\label{eq:mahalanobis-distance}
D_{i,k}^{2} =  (h(x_k,\nu_k)-h(\mu_i,0))^T R_k^{-1} (h(x_k,\nu_k)-h(\mu_i,0)).
\end{equation}

\begin{figure*}
  \centering
  \subfigure[A candidate policy is generated for each mode and we pick the policy with maximum expected gain.]{\includegraphics[width=1.5in]{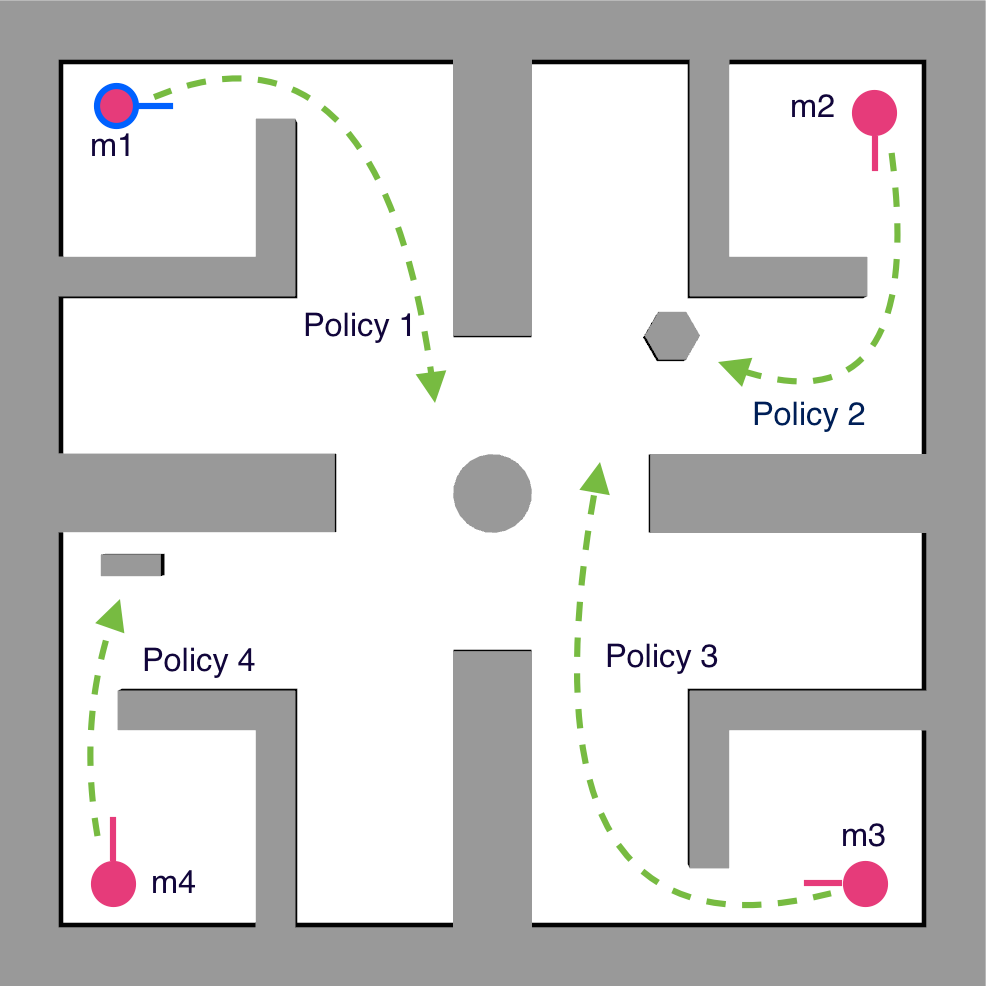}}
  \hspace{0.3in}
  \subfigure[Policy 3 is chosen and executed, leading all the hypothesis out of the different rooms. Mode $m_4$ expects to see a landmark outside the door which the robot does not see.]{\includegraphics[width=1.5in]{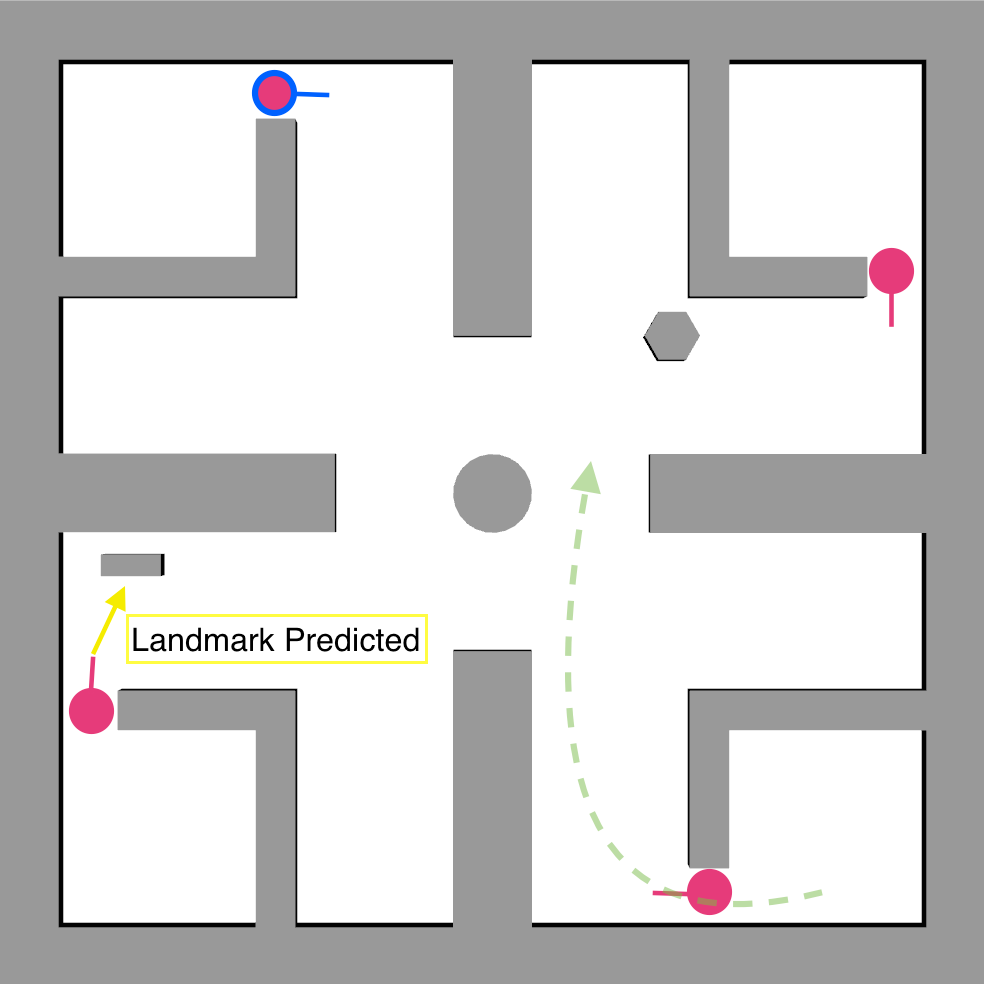}}
  \hspace{0.3in}
  \subfigure[Mode $m_4$ is rejected and a new set of candidate policies is computed.]{\includegraphics[width=1.51in]{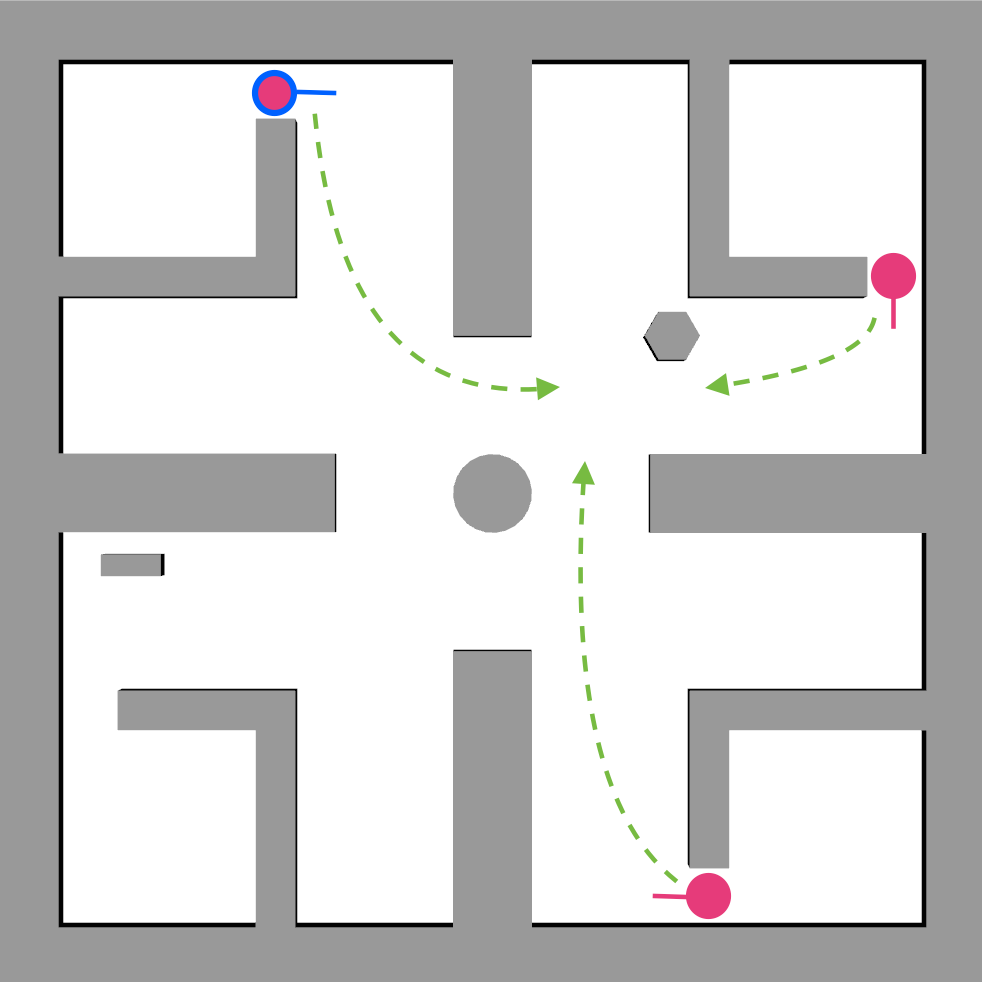}}
  \caption{Key steps in the Receding Horizon Control methodology for a multi-modal belief.}
  \label{fig:planner-rhc-method}
\end{figure*}

\begin{algorithm}[t]
\caption{Generating Multi-Modal Belief State} \label{alg:sampling-gmm}

  Input: $\Sigma_0$ (Initial Covariance) \\
  
  $\{\mu \} =$ Sample valid states uniformly; \\ 
  
  \For{$\mu_i \in \{\mu \}$}
  {
  	$m_i =$ Compose belief mode with mean $\mu_i$ and covariance $\Sigma_0$ ;\\
  	
	Add mode $m_i$ to belief $b$ ;\\
  }
  
  Assign uniform weight to all modes ; \\
  
  \While{ not converged to fixed number of modes}
  {
  	
  	Update mode weights in belief $b$;  \\ 
  	
    Remove modes with weights lower than threshold $\delta_w$; \\
  }
  
 \Return $b$;
\end{algorithm}

Depending on the state of the robot, individual hypotheses and data association results, we might have several cases.

\begin{enumerate}
\item The number of real landmarks observed $n_z$ could be equal, less or more than the number of predicted observations $n_{h_i}$ where $h_i$ is the predicted observation vector for the $i'th$ mode. 
\item The number of associated real and predicted observations $n_{z\cap{h} } $ could be equal or less than $n_z$, $n_h$ or both. 
\end{enumerate}

Thus, to update the weight of each hypothesis, we factor in the above information. First we estimate the Mahalanobis distance $D_{i,k}$ between the predicted and observed landmarks that are matched by the data association module. Then we multiply this weight by a factor $\gamma$, which models the effect of duration for which the robot and a hypothesis predict to see different information. The entire weight update algorithm is described in algorithm \ref{alg:gmm-weight-update}. After each weight update step, we remove modes with negligible contribution to the belief i.e. when $w_{i,k+1} < \delta_w$ where $\delta_w$ is a user defined parameter for minimum weight threshold (1\% of the total weight in our simulations).

\begin{algorithm}[h!]
\caption{GMM Weight Update} \label{alg:gmm-weight-update}

Input: $w_{i,k}, \mu_{i,k+1}$ \\
Output: $w_{i,k+1}$; \\

$z_{k+1} = $  Get sensor observations ;\\

$n_z = $ Number of landmarks observed in $z_{k+1}$; \\

$n_{z\cap{h}} =$ Data association between $h(\mu_{i,k+1}, 0)$ and $z_{k+1}$ ;\\


$w^{'}_{i,k+1} =$ Update and normalize weight according to likelihood function ; \\

\If{$n_{h} \neq n_z  ~~ ||~~ n_h \neq n_{z\cap{h} } $}
{
  $ \alpha = max( 1+n_z-n_{z\cap h} , 1+n_h-n_{z\cap h} )$ ;\\
  $ \beta = \beta + \delta t $ ;\\
  $ \gamma = e^{-0.0001\alpha\beta}$ ;\\
}
\Else
{
  $ \beta = \beta - 1.0 $ ;\\
  \If{$ \beta < 0$}
  {
    $ \beta = 0$; \\
  }
}

$w_{i,k+1}  = w^{'}_{i,k+1} \gamma $ ;\\

\Return $w_{i,k+1}$;
\end{algorithm}

\subsection{Multi-Modal Motion Planner (M3P)}\label{sec:mmpolicygen}

The task of the non-Gaussian planner is to generate a sequence of control actions such that the belief state converges to a uni-modal Gaussian. We call our planner \textbf{M3P} i.e. Multi-Modal Motion Planner. 
Algorithm \ref{alg:m3p} describes the planner's key steps and Fig. \ref{fig:planner-rhc-method} describes the basic functionality of the planner in a sample scenario.\\

\begin{algorithm}[t]
\caption{M3P: Multi-Modal Motion Planner} \label{alg:m3p}
  Input: $b$ (Belief) \\
  
  \While{$b \neq \mathcal{N}(\mu, \Sigma)$}
  {
    $\Pi$ = Generate candidate policy for each belief mode ;\\
    
    $\pi$ = Pick policy from $\Pi$ with maximum expected information gain ; \\
    
    \ForAll{$u \in \pi$}
    {
      $b$ = Apply action $u$ and update belief; \\
      
      \If{Change in number of modes $||$ Expect a belief mode to violate constraints}
      {
	break; \\
      }
    }
  }
  
  \Return $b$;
\end{algorithm}

The key steps in the M3P planner are steps 3 and 4 in Algorithm 3 that generate the candidate policies for the different modes, and then choose one of the policies based on the expected information gain. In the following section, we give an uniqueness graph (node) based instantiation of the planner which allows us to recover from kidnapped robot scenarios.


\subsection{Node based Multi-Modal Motion Planner (NBM3P)}
Here we give a particular instantiation of the two key steps required to enable the M3P planner; 1) generating the candidate policies and, 2) picking the optimal policy.

\subsubsection{Generating a Set of Candidate Policies}\label{sec:uniqueness-graph}

In a multi-modal scenario no-one action based on a belief mode can be said to be the best action. The best action for one mode may or not be good for the other modes. Thus, in a multi-modal scenario, we claim that the best action to take is one that guides the robot without collision through a path that results in information gain such that the belief converges to a uni-modal Gaussian pdf. We reduce the infinite set of possible control actions to a discrete number of policies by picking a target state for each mode to go to and
generate control actions that guide each mode to its corresponding target. \\

\textit{Picking the target state for a mode}: We introduce the concept of a uniqueness graph $G_u$ (computed offline) that allows us to pick target states for each belief mode. The uniqueness graph $G_u$ is constructed using information from the FIRM graph. Each FIRM node is added as a node in $G_u$ (only $x,y,\theta$). Once a node is added, we calculate what information the state represented by that node can observe in the environment. Using this information we add an edge $E_{\alpha \beta}$ (undirected) between two nodes $v_{\alpha}$ and $v_{\beta}$ if both nodes see similar information. In our landmark based observation model, each landmark has an ID. Thus if both nodes observe the same ID or IDs then an edge is added and the weight of the edge is equal to the number of similar landmarks observed. The edge weight gives an indication of the similarity (or conversely uniqueness) in the information observed. If the weight of an edge is higher, it means the states represented by the vertices of that edge are more likely to observe similar information. To find the target for a belief mode $m_i$, we first choose the set of nodes $N_i$ which belong to the neighborhood of $\mu_i$ (which is the mean of mode $m_i$). Then, we find the node $v^{t}_i \in N_i$ which observes information that is least similar in appearance to that observed by nodes in the neighborhood $N_j$ of mode $m_j$ where $j \neq i$. We are trying to solve the optimization problem, \\



\begin{equation}
(v^{t}_1, v^{t}_2 \hdots v^{t}_n) = \operatorname*{arg\,min}_{v_1, v_2, \hdots, v_n} (h(v_1,0) \cap h(v_2,0) \cap \hdots \cap h(v_n,0)).
\end{equation}

To do this, first we calculate the total weight of the outgoing edges from every node $v \in N_i$ to nodes in all other neighborhoods $N_j$ where $j \neq i$. The node which has the smallest outgoing edge weight, is the target candidate. Algorithm \ref{alg:find-target-node} describes in detail the steps involved. 

\begin{algorithm}[h!]
\caption{Finding the target for a mode in a multi-modal belief} \label{alg:find-target-node}
{
  Input: $m_{i,k}$ , $G_u$ ; // belief mode, uniqueness graph \\ 
  Output: $v^t_i$; // target node \\ 

  \ForAll{$m_{l,k} \in M_k$}
  {
	  $N_l =$ Find nodes in $G_u$ within neighborhood of radius $R$ centered at $\mu_{l,k}$ ;\\
  }

  $minWeight$ = Arbitrarily large value; \\
  
  $v^t_i = -1$ ; \\

  \ForAll{$ v \in$ Neighborhood $N_i$ of mode $m_{i,k}$}
  {
    
    \For{$ N_j \in$ set of all neighborhoods of the belief modes and $j \neq i $}
    {
      $w = 0$ ;\\
      \ForAll{$e \in$ Edges connected to $v$}
      {
	      \ForAll{$p \in N_j $}
	      {
		      \If{$p$ is a target of edge $e$}
		      {
			      $w += \mathtt{edgeWeight}(e)$; \\
		      }

	      }
      }
     
    }
     \If{$w \leq minWeight$}
      {
	      $minWeight = w$; \\
	      $v^t_i = v$ ;\\
      }
  }
  
  \Return $v^t_i$; \\
}
\end{algorithm}

\textit{Connecting a mode to its target}: Once we have picked the targets corresponding to each mode, we need to find the control action that can take the mode from its current state to the target state. We generate the open loop control sequence that takes each mode to its target using the RRT* planner. RRT* is chosen because it is computationally cheap and can incorporate the system's kinodynamical constraints.

\subsubsection{Picking the Optimal Policy}

Once we have generated the set of candidate policies. We need to evaluate the expected information gain from each policy and pick the optimal policy that maximizes this information gain. We model this information gain as the discrete change in the number of modes. This implies that instead of doing Monte Carlo simulations with noisy observations and motion to generate all possible belief trajectories, we can simulate the most-likely belief trajectory. This helps reduce the computational burden significantly. We know that one policy may or may not be good for all the modes i.e. a policy based on one mode may lead to collision for the other modes. Therefore, we need a way of penalizing a candidate policy if it results in collision. We introduce a penalty $c_{fail} / k$ where $c_{fail}$ is a fixed value ($10^6$) and $k$ is the step during execution at which the collision takes place.
Thus, policies which result in a collision much further down are penalized less compared to policies that result in immediate collision.
The steps to calculate the expected information gain for a candidate policy are:

\begin{enumerate}
 \item Pick a candidate policy $\pi_i$ corresponding to mode $\mu_i$
 \item Assume that the robot is actually at one of the modes. 
 \item Execute the policy and propagate all the modes and measure the information gain (the reduction in the number of modes).
 \item Repeat steps 2-3 for each mode and sum the weighted information gain weighted by the mode weights.
\end{enumerate}

We repeat the above steps for all policies in the set $\Pi$ and pick the policy with maximum weighted information gain.

Algorithm \ref{alg:OL-RRT} shows how we generate the optimal policy $\pi$. Two key functions to note here are:

\begin{enumerate}
 \item \textit{findTargetState($\bar{b}^{i}_{k}$)} : Finds a FIRM node in the neighborhood of mode $\bar{b}^{i}_{k}$ at time $k$ such that by going to this node, the robot can gain the maximum possible information. Algorithm \ref{alg:find-target-node} shows how this function works.
 
 \item \textit{calculateInformationGain($\pi_i,\bar{b}^{i}_{k}$)} : Given a control sequence $\pi_i$ and mode $\bar{b}^{i}_{k}$, it calculates the expected information gain. Algorithm \ref{alg:expected-information-gain} shows how the expected information gain is calculated for the candidate policies.
 
\end{enumerate}

\begin{algorithm}[t]
\caption{Calculating the Optimal Policy $k$} \label{alg:OL-RRT}

  \ForAll{$m_{i,k} \in M_k $}
  {
    $v^t_{i,k} =$ Find target state for mode $(m_{i,k})$ ; \\
    
    $\pi =$ Find open loop controls from $\mu_{i,k}$ to $v^t_{i,k}$ using RRT* ;\\
    
    $\Pi.\mathtt{append}(\pi)$; \\
  }
  \ForAll{$\pi_j \in \Pi$}
  {
    \ForAll{$m_{i,k} \in M_k$}
    {
      $\Delta I_{ij} =$ Calculate expected information gain for policy $\pi_j$ assuming robot is at $\mu_{i,k}$; 
    }
    
    $\delta I_{j} = w_j \sum\limits_{j=1}^n \Delta I_{ij} $ ; \\
  }
  
  $\pi^{*} =$ Pick policy from $\Pi$ that has maximum gain $\delta I$ ;\\

  \Return $\pi^{*}$;
\end{algorithm}

\begin{algorithm}[h!]
\caption{Calculating Expected Information Gain} \label{alg:expected-information-gain}
{
  Input: $\pi_j, m_{i,k}$ \\ 
  \vspace{0.1cm}
  Output: $\Delta I_{ij}$ \\ 
  
   $x_0 = \mu_{i,k}$ ; \\
  
   $n_{0}$ = Current number of belief modes;\\
   $\Delta I_{ij} = 0 $ ; \\
%
  
  \ForAll{$u_k \in \pi_i $}
  {
      $x_{k+1} = f(x_{k},u_{k},0)$ ; \\
      
      $z_{k+1} = h(x_{k+1},0)$ ; \\
      
      
      Propagate individual Kalman filters with control $u_k$ and most likely observation $z_{k+1}$ ;\\
                
      
      \If{ Any belief mode is in collision configuration}
      {
	 	$\Delta I_{ij} = \Delta I_{ij} - \frac{c_{fail}}{k} $;\\
		$\mathtt{break}$; \\
      }
      
  }
  
  $n_{T}$ = Number of belief modes after simulation;\\
  
  $\Delta I_{ij} = \Delta I_{ij} + n_{T} - n_{0}$ ;
  
  \Return  $\Delta I_{ij}$;
  
}
\end{algorithm}

%
%
%
%
%
%
%
%
%
%
%
%
%
%
%

\subsection{Analysis}
In this section, we show that the basic receding horizon planner M3P  will guarantee that an initial multi-modal belief is driven into a unimodal belief in finite time. First, we make the following assumptions:
\begin{assumption}
Given a multi-modal belief $b_k = \sum_i w_{i,k}m_{i,k}$, for every mode $m_{i,k}$, there exists a disambiguating planner $\mu_i(.)$ in the sense that if the robot was really in mode $i$, the planner's actions would confirm that the robot was at mode $i$.
\end{assumption}
\begin{assumption}
The map does not change during the execution of the planner.
\end{assumption}
\begin{proposition}
Under Assumptions 1 and 2, given any initial multi-modal belief $b_0 = \sum_i w_{i,0}m_{i,0}$, the receding horizon planner M3P drives the belief process into a unimodal belief $b_T = m_T \approx \mathcal{N}(\mu_T, \Sigma_T)$ in some finite time $T$.
\end{proposition}
\begin{IEEEproof}
Suppose that the robot is at the initial belief $b_0$. Suppose we choose the plan $\mu_{i^*}$ that results in the most information gain as required by the M3P planner. The plan $\mu_{i^*}$ can be applied to all the modes at least for some finite period of time in the future,  since if it cannot be applied, then we immediately know that the robot is not at mode $i^*$ and thus, there is a disambiguation whereby mode $i^*$ is discarded.\\
Once the plan $\mu_{i^*}$ is executed, there are only 2 possibilities:\\
1) The robot is able to execute the entire plan $\mu_{i^*}$ till the end, or
2) the plan becomes infeasible at some point of its execution. \\
In case 1, due to Assumption 1, we will know for sure that the robot was at mode $i^*$ and the belief collapses into a unimodal belief thereby proving the result. In case 2, due to Assumption 2, we know that the robot could not have started at mode $i^*$ and thus, the number of modes is reduced by at least one. After this  disambiguation, we restart the process as before and we are assured that atleast one of the modes is going to be disambiguated and so on. Thus, it follows given that we had a finite number of modes to start with, the belief eventually converges to a unimodal belief. Further, since each of the disambiguation epochs takes finite time, a finite number of such epochs also takes a finite time, thereby proving the result.\\
\end{IEEEproof}
The above result shows that the basic M3P algorithm will stabilize the belief process to a unimodal belief under Assumptions 1 and 2. In order to show that a particular instantiation of the planner, such as the target node based planner NBM3P detailed above, stabilizes the belief to a single mode, we need to show that the particular planner satisfies Assumption 1. It can be seen that under NBM3P, by design, the robot reaches a unique disambiguating location if it started in a particular mode, and thus, Assumption 1 is satisfied. Hence, this leads to the following corollary.
\begin{corollary}
The Node based receding horizon planner, NBM3P, drives any initial multi-modal belief into a unimodal belief in finite time, under Assumption 2.
\end{corollary}

\begin{remark}
The above results hold if we choose the disambiguating policy at random, and not as the optimal one in the sense of the number of modes disambiguated. However, in practice we see that the optimal disambiguating policy gives us better performance.
\end{remark}

\section{Simulation Results}\label{sec:simulations}

We present simulation results for a 2D robot. The simulations represent a motion planning scenario wherein the robot is tasked to go from a start to a goal location in an environment where there is symmetry. We initially rely on FIRM \cite{Ali14-IJRR} to provide a feedback policy. However, en-route to the goal, the robot is kidnapped to an unknown location and it cannot localize. Thus, it relies on the non-Gaussian planner NBM3P described previously to localize. Once the robot is localized, its new belief is connected to the existing FIRM graph and we find a new feedback policy to complete the task. We abstract the geometry/appearance of the environment by using passive visual beacons that have IDs associated to them. We place multiple landmarks with the same IDs in different locations, thus creating the `illusion' of multiple-data associations. All simulations were carried out on a Macbook Pro laptop with an Intel Core-i5 2.6GHz CPU and 8GB of RAM running Ubuntu 14.04, we use the Open Motion Planning Library \cite{ompl-library} as the software back-end. \\
\textit{A supplementary video is provided \cite{youtube-video-m3p} that clearly depicts every stage of the simulations with a robot operating in a virtual environment.}
The key steps in the simulation are:

\begin{enumerate}
 \item Generate an initial policy from start to goal. 
 \item Keep tracking innovation, large change causes robot to detect lost state.
 \item Switch from Gaussian to a GMM representation of the underlying belief.
 \item Generate information gathering actions using M3P that are able to localize the state.
 \item Switch back to uni-modal Gaussian and connect the belief to existing FIRM roadmap.
 \item Find new policy to goal.
\end{enumerate}

\subsection{Motion Model}\label{sec:experiments-motion-model}

We simulate a 2D ground robot, the kinematics of which are represented by a unicycle.

\begin{align}\label{eq:unicycle-motion-model}
\!\!\!\!\!x_{k+1}& \!=\! f(x_k,u_k,w_k) \!=\!
\left(\!
  \begin{array}{c}
    \mathsf{x}_{k}+(V_k + n_v)\delta t\cos\theta_k \\
    \mathsf{y}_{k}+(V_k + n_v)\delta t\sin\theta_k \\
    \mathsf{\theta}_{k}+(\omega_k + n_{\omega})\delta t
  \end{array}\!\right)\!,
\end{align}
where $ x_k = (\mathsf{x}_k, \mathsf{y}_k, \mathsf{\theta}_k)^T $ describes the robot state (position and yaw angle). $ u_k = (V_k,\omega_k)^T $ is the control vector consisting of linear velocity $ V_k $ and angular velocity $ \omega_k $. We denote the process noise vector by $ w_k=(n_v,n_{\omega})^T\sim\mathcal{N}(0,\mathbf{Q}_k) $.


\subsection{Observation Model}\label{sec:experiments-observation-model}

Our observation model is based on passive beacons/landmarks which can be pinged to measure their relative range and an associated ID tag. This is parallel to a physical implementation using a monocular camera and passive Augmented Reality (AR) beacons that can be detected by the camera. We use the visual AR model to simplify the data association problem. \\

Let the location of the $i$-th landmark be denoted by ${^i}\mathbf{L}$. The displacement vector ${^i}\mathbf{d}$ from the robot to ${^i}\mathbf{L}$ is given by ${^i}\mathbf{d}=[{^i}d_{x}, {^i}d_{y}]^T:={^i}\mathbf{L}-\mathbf{p}$, where $\mathbf{p}=[\mathsf{x},\mathsf{y}]^T$ is the position of the robot. Therefore, the observation ${^i}z$ of the $i$-th landmark can be modeled as,

\begin{align}
{^i}z={^i}h(x,{^i}v)=[\|{^i}\mathbf{d}\|,\text{atan2}({^i}d_{y},{^i}d_{x})-\theta]^T+{^i}v,
\end{align}

The observation noise is zero-mean Gaussian  such that $ {^i}v\sim\mathcal {N}(\mathbf{0},{^i}\mathbf{R}) $ where ${^i}\mathbf{R}=\text{diag}((\eta_r\|{^i}\mathbf{d}\|+\sigma^r_b)^2,(\eta_{\theta}\|{^i}\mathbf{d}\|+\sigma^{\theta}_b)^2)$. 
The quality of sensor reading decreases as the robot gets farther from the landmarks. 
The parameters $\eta_r$ and $\eta_{\theta}$ determine this dependency, and $\sigma_b^r$ and $\sigma_b^{\theta}$ are the bias standard deviations. The robot observes only those landmarks that fall within its sensor range $r_{sensor}$. 

\subsection{Scenario}\label{subsec:6corridor-world}

\begin{figure}[ht!]
 \centering
 \includegraphics[width=3.5in]{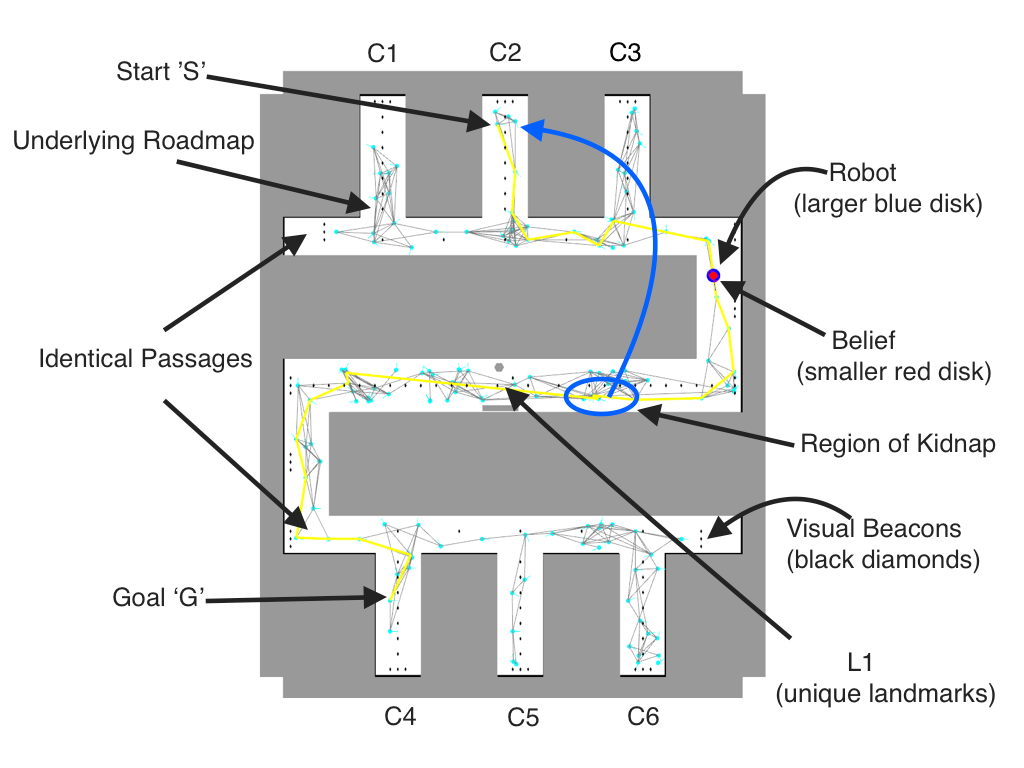}
 \caption{The environment depicting obstacles (gray) and free space (white). The cyan disks represent the nodes in the FIRM roadmap. The robot is en-route to the goal before kidnapping and the initial plan under FIRM is demarcated by yellow. The blue ellipse marks the region where the robot will be kidnapped (not known to robot) and the final location after kidnapping (in \textit{C2}) is marked by the blue arrow (begins inside the ellipse).}
 \label{fig:6corridorworld}
\end{figure}

\begin{figure}[ht!]
  \centering
  \subfigure[The robot just before kidnapping (robot is not aware of impending kidnapping).]{\includegraphics[height=2in]{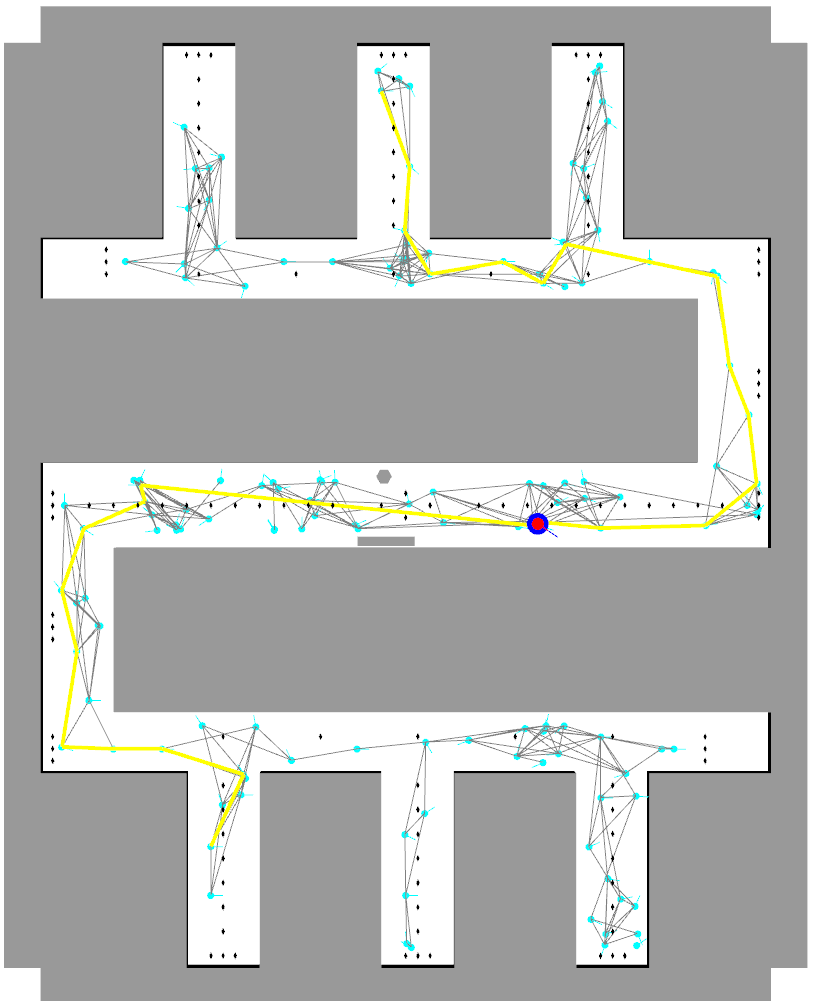}}
    \hspace{0.1in}
  \subfigure[Robot and multi-modal belief just after kidnapping. Each red (smaller) disk represents a belief mode.]{\includegraphics[height=2in]{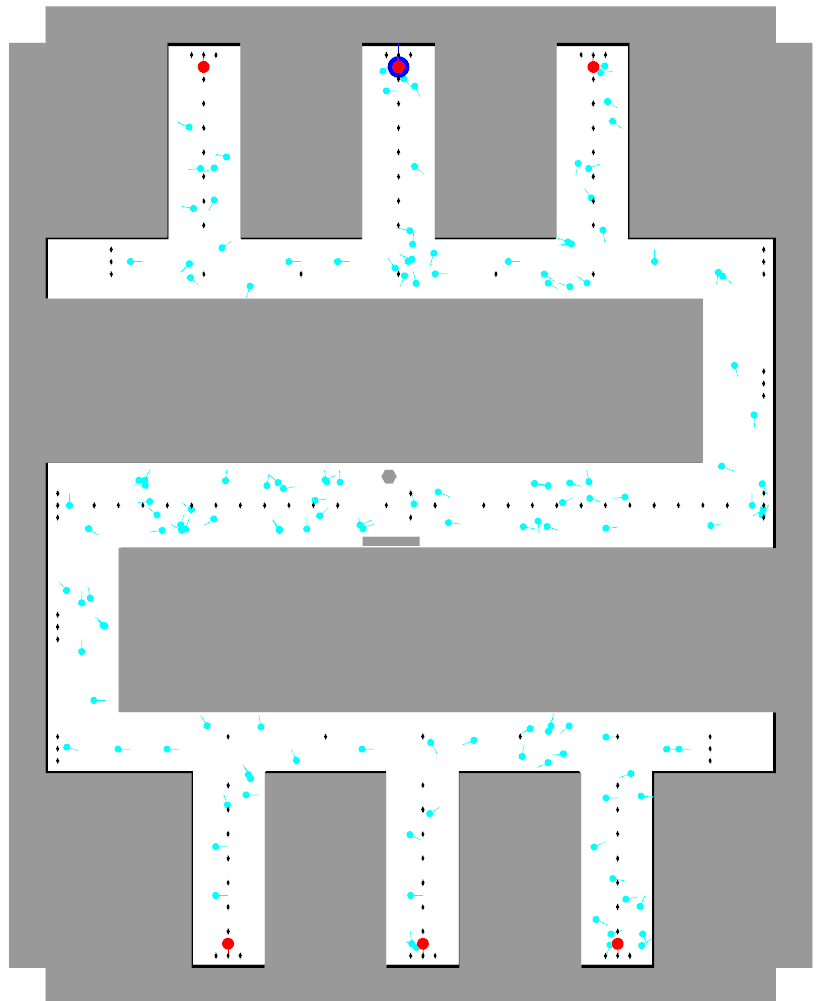}}
  \caption{Before and after kidnapping.}
  \label{fig:6corridorworld-kidnapping}
\end{figure}

The environment as shown in Fig. \ref{fig:6corridorworld} represents a warehouse floor where 6 corridors \textit{C1-6} appear exactly alike. These corridors open into two passages which also appear alike. There are unique landmarks at location \textit{L1}. The robot is put at the start location \textit{S} and tasked to reach goal location \textit{G} as shown. Using FIRM, we generate the feedback policy from \textit{S} to \textit{G}. The blue ellipse and arrow mark the region where the robot will be kidnapped and where it will be displaced to respectively
First, we generate an initial feedback policy from \textit{S} to \textit{G}. When the robot is en-route to \textit{G} (Fig. \ref{fig:6corridorworld-kidnapping}a) it is kidnapped and placed in corridor \textit{C2} as shown in Fig. \ref{fig:6corridorworld-kidnapping}b. A new multi-modal belief is generated which took $51.23$s to compute.


\begin{figure}
  \centering
  \subfigure[Initial multi-modal belief and candidate policies at time $t_1$.]{\includegraphics[height=2.2in]{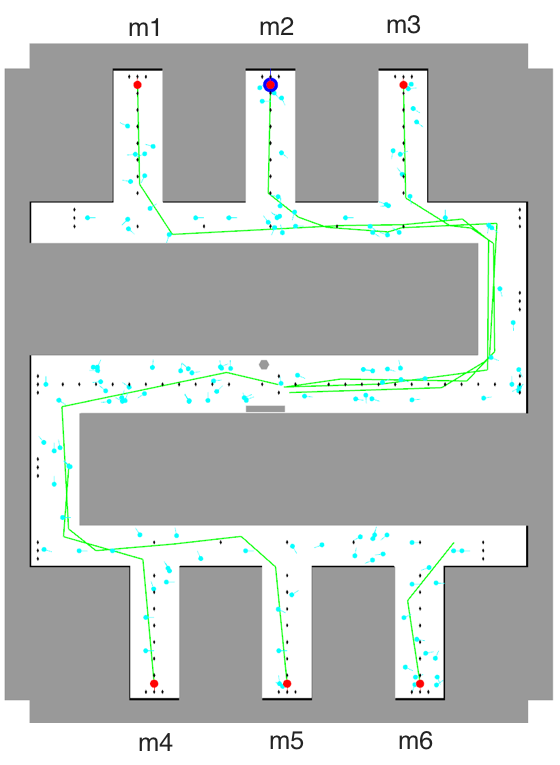}}
      \hspace{0.1in}
  \subfigure[Modes $m_3$ and $m_4$ are rejected at $t_2$ as they expect to see a corner to their left on exiting their respective corridors.]{\includegraphics[height=2in]{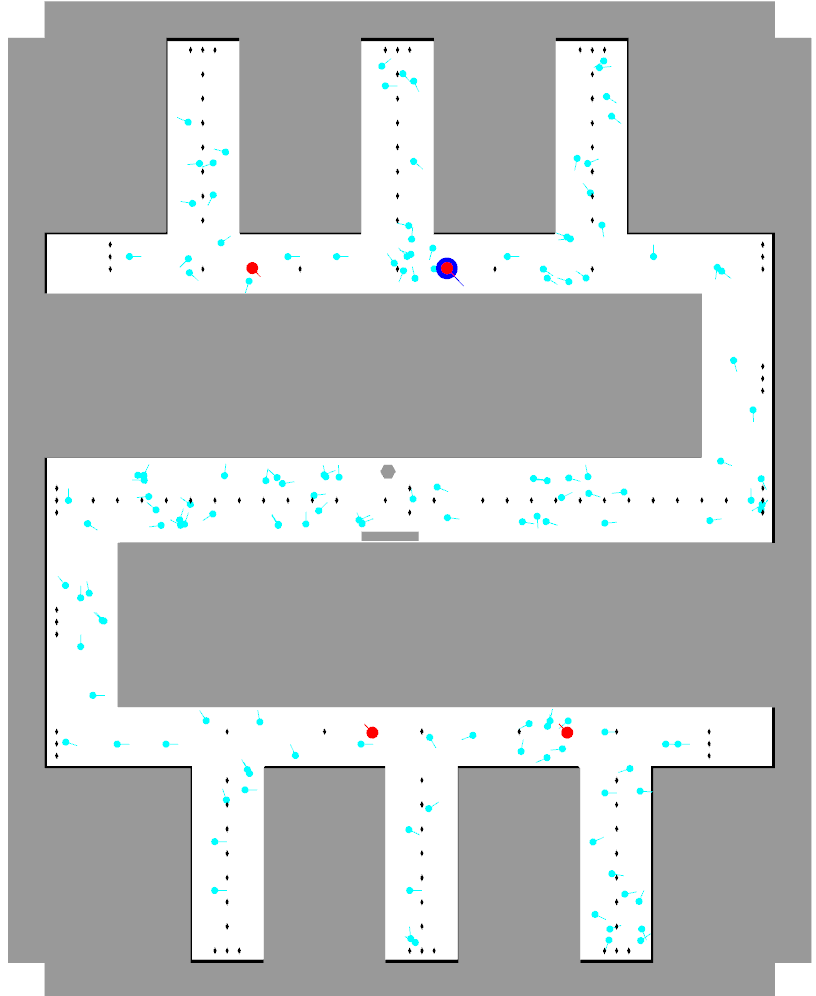}}\quad
  \subfigure[Following a new open loop policy at $t_3$. This leads modes $m_1$ and $m_6$ to be rejected.]{\includegraphics[height=2in]{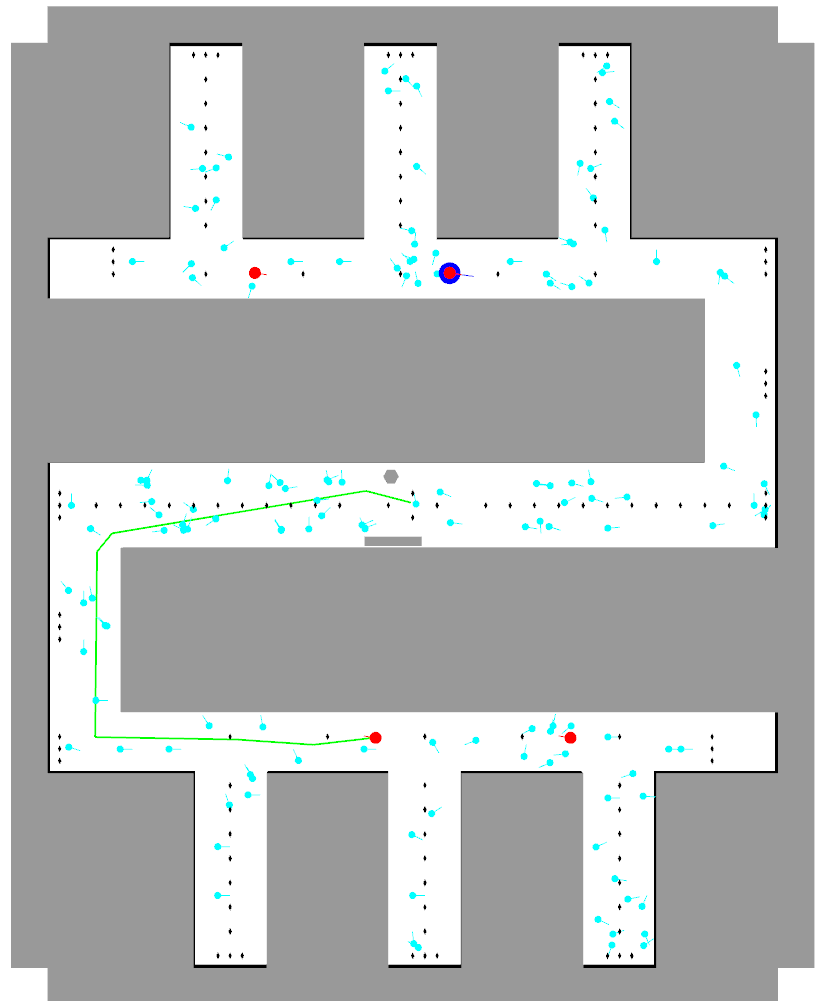}}
   \hspace{0.1in}
  \subfigure[Modes $m_2$, $m_5$ and the robot approach the distinguishing landmark \textit{L1} at $t_4$.]{\includegraphics[height=2in]{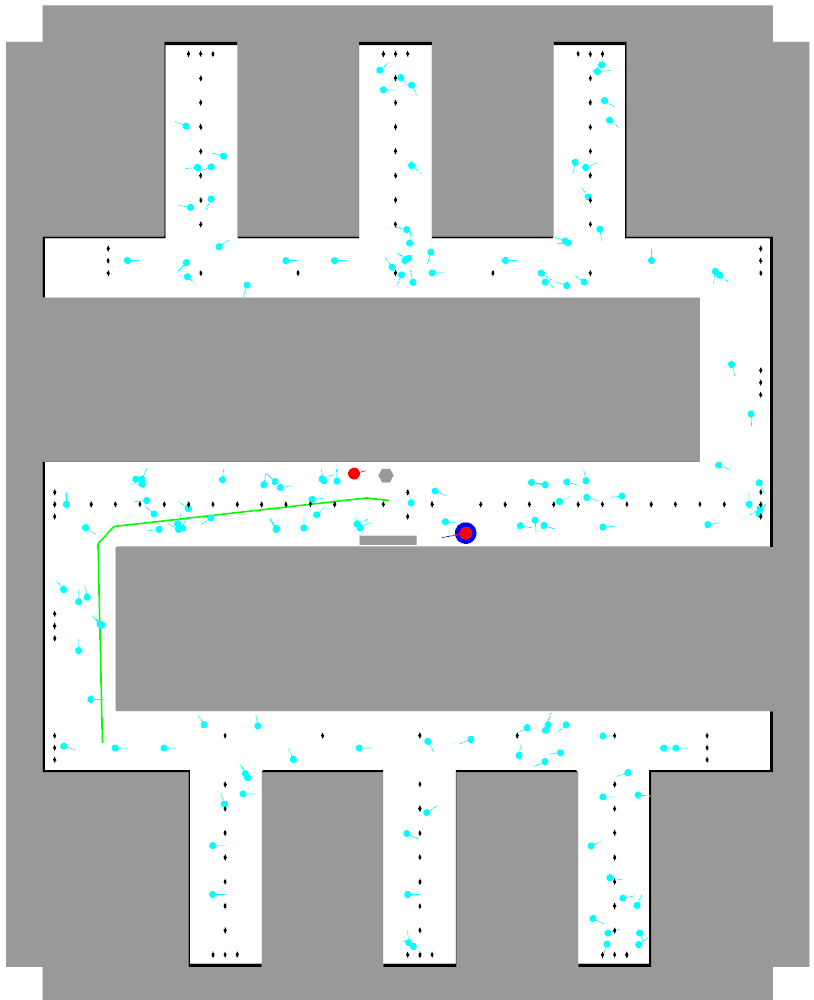}}
  \caption{Candidate policies and belief modes during recovery from kidnapping.}
  \label{fig:6corridorworld-rrt}
\end{figure}

We analyze the recovery from kidnapping at four different time steps:

\begin{enumerate}
 \item $t_1$: Fig. \ref{fig:6corridorworld-rrt}a shows the initial candidate policies after the new belief is sampled. The policy was computed in $58.76$s.
 
 \item $t_2$: Fig. \ref{fig:6corridorworld-rrt}a shows that on exiting the corridors, the modes $m_3$ and $m_4$ that originated in $C3$ and $C4$ expect to see a corner to their left and are rejected as this is not observed by the robot. A new policy is then computed which took $12.48$s.
 
 \item $t_3$: In Fig. \ref{fig:6corridorworld-rrt}c we see the new candidate policy. This eventually leads the robot to move forward which results in the modes $m_1$ and $m_6$ that exit $C1$ and $C6$ to be rejected as the robot expects to see a corner. Once these two modes are rejected a new policy is computed which took $3.7$s.
 
 \item $t_4$: Fig. \ref{fig:6corridorworld-rrt}d shows the final two remaining modes approaching the unique landmark which eventually leads to a uni-modal belief localized to the true state.
\end{enumerate}

\begin{figure}[ht!]
  \centering
  \subfigure[Belief is localized to the true state. This new belief is added to existing roadmap.]{\includegraphics[height=2in]{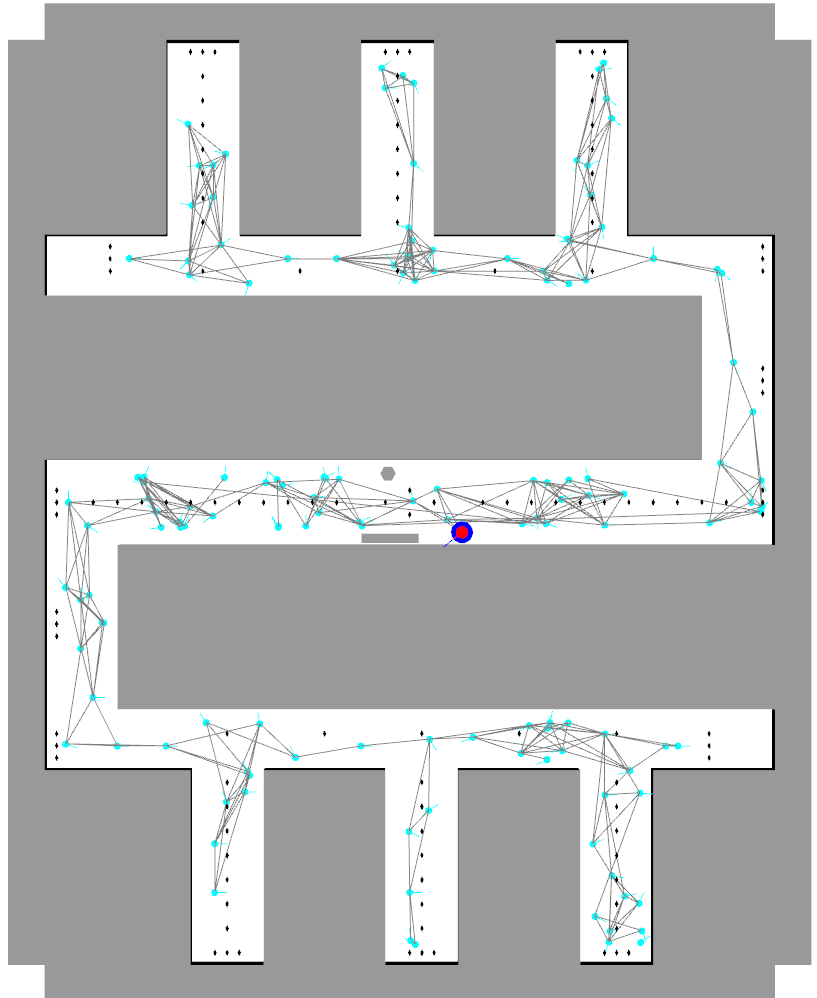}}
  \subfigure[Task completed, the robot successfully reached the goal location.]{\includegraphics[height=2in]{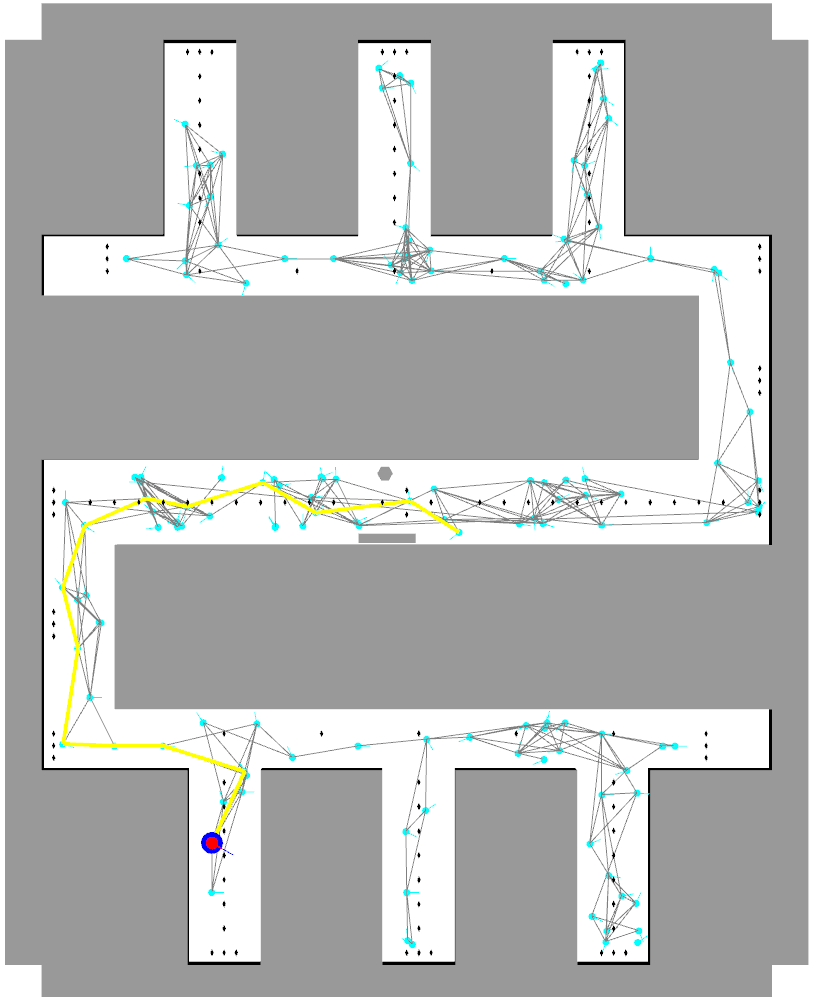}}
  \caption{Recovery from kidnapping}
  \label{fig:6corridorworld-localized}
\end{figure}

Fig. \ref{fig:6corridorworld-localized}a shows the belief localized, once this belief is added to the existing FIRM roadmap, a new policy guides the robot to the goal (Fig. \ref{fig:6corridorworld-localized}b). 

\subsection{Discussion}
 
It is seen that a highly intuitive behavior emerges which guides the robot to seek disambiguating information such that it can sequentially reject the incorrect hypothesis about its state. The open-loop control actions are regenerated every time a Gaussian mode is rejected or a constraint violation is foreseen. The time to re-plan reduces drastically as the number of modes reduce ($O(n^2)$). Thus, the first few actions are the hardest which is to be expected as we start off with a large number of hypotheses. Finally, the planner is able to localize the robot safely. 

\section{Conclusion}

In this work, we studied the problem of motion planning for a mobile robot when the underlying belief state is non-Gaussian in nature. A non-Gaussian belief can result from uncertain data associations or due to unknown initial conditions among others. Our main contribution in this work is a planner M3P that generates a sequentially disambiguating policy, which leads the belief to converge to a uni-modal Gaussian. We are able to show in simulation that the robot is able to recover from a kidnapped state and execute its task in environments that present multiple uncertain data associations. Our next step would be to demonstrate the same on a physical system and prove its robustness in real-life situations.
%
%

\bibliographystyle{plainnat}
\bibliography{references}

\end{document}